\documentclass[runningheads]{llncs}
\usepackage{graphicx}
\usepackage{cite}
\usepackage{amsmath,amssymb,amsfonts}
\usepackage{algorithmic}
\usepackage{graphicx}
\usepackage{textcomp}
\usepackage{booktabs}
\usepackage{xcolor}
\usepackage{comment}

\begin{document}
\pdfoutput=1
\title{Tag-based Semantic Features \\ for Scene Image Classification
}
%
%

\author{Chiranjibi Sitaula\inst{1}
\and
Yong Xiang\inst{1}
\and
Anish Basnet\inst{2}
\and
Sunil Aryal\inst{1}
\and
Xuequan Lu\inst{1}
}

\authorrunning{C. Sitaula et al.}
%

\institute{School of Information Technology, Deakin University, Geelong, Australia\\ 
\email{\{csitaul, yong.xiang, sunil.aryal, xuequan.lu\}@deakin.edu.au} \and
Ambition College, Kathmandu, Nepal\\
\email{anishbasnetworld@gmail.com}
}

%
\maketitle              
\begin{abstract}
The existing image feature extraction methods are primarily based on the content and structure information of images, and rarely consider the contextual semantic information. 
Regarding some types of images such as scenes and objects, the annotations and descriptions of them available on the web may provide reliable contextual semantic information for feature extraction. In this paper, we introduce novel semantic features of an image based on the annotations and descriptions of its similar images available on the web. Specifically, we propose a new method which consists of two consecutive steps to extract our semantic features. For each image in the training set, we initially search the top $k$ most similar images from the internet and extract their annotations/descriptions (e.g., tags or keywords). The annotation information is employed to design a filter bank for each image category and generate filter words (codebook). 
Finally, each image is represented by the histogram of the occurrences of filter words in all categories. We evaluate the performance of the proposed features in scene image classification on three commonly-used scene image datasets (i.e., MIT-67, Scene15 and Event8). Our method typically produces a lower feature dimension than existing feature extraction methods. Experimental results show that the proposed features generate better classification accuracies than vision based and tag based features, and comparable results to deep learning based features. 

\end{abstract}
\section{Introduction}
\begin{figure}[b]
\begin{center}
 \includegraphics[width=0.45\textwidth, height=30mm,keepaspectratio]{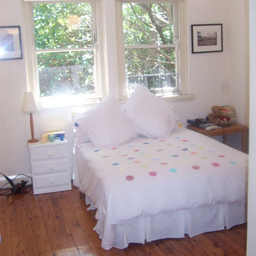}
\hspace{15pt}
 \includegraphics[width=0.45\textwidth, height=30mm,keepaspectratio]{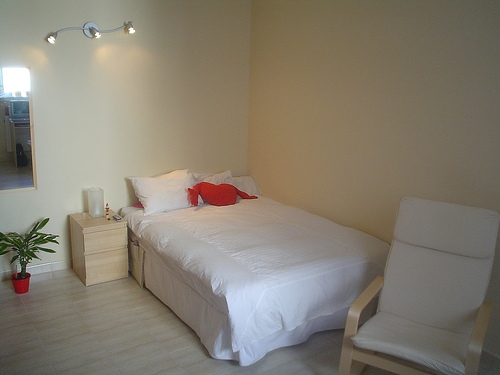}
   \caption{The images of bed room (left) and hospital room (right) look similar. }
  \label{fig:1}
 \end{center}
  \end{figure}

In computer vision, features of images are extracted primarily from the content and structure of images. They rely on information such as color, texture, shapes, and parts. Though these features are shown to work reasonably well in many image processing tasks \cite{zeglazi_sift_2016,oliva2005gist,oliva_modeling_2001,dalal2005histograms,wu_centrist:_2011,xiao_mcentrist:_2014, margolin2014otc}, the involved information may be insufficient to distinguish ambiguous images, e.g., classifying images with  inter-class similarity. Fig.~\ref{fig:1} shows two images which look very similar but they belong to two different categories (hospital room and bedroom). 

Contextual information is useful to distinguish such ambiguous images\cite{sitaula2019indoor,li2010object}. Contextual information about the image
can be often obtained from its annotations or descriptions. Though it is impossible to have descriptions or annotations for all images, such information can be extracted from the web for certain types of images like objects and scenes. For the extraction of such contextual features,  
we use the well-known search engine Yandex\footnote{https://www.yandex.com/images/} for two reasons: firstly, visually similar images of image categories such as wine cellar are not meaningful while using the Google search engine; secondly, we notice that the searched similar images of the input image usually belong to the same category.
  
Recently few prior works\cite{zhang2017image,wang2019task} have been proposed for scene image recognition using tag-based features. These methods suffer from the following major issues.
\begin{itemize}
    \item The task-generic filter banks\cite{wang2019task} lack context-based information for the images such as scene images, owing to their generality based on the help of pre-defined labels of ImageNet\cite{deng_imagenet:_2009} and Places \cite{zhou2016places} dataset.
    \item The existing tag-based method\cite{zhang2017image} does not design filter banks, thus resulting in high dimensional features with noticeable outliers.
\end{itemize}

As such, these state-of-the-art techniques yield limited classification accuracy. In this paper, we introduce new semantic features of an image based on the annotation/description tags of similar images available on the web. We design two consecutive steps to extract our proposed features after we select top $k$ most similar images of each image under the dataset using Yandex and extract annotation/description tags corresponding to those images. 
At first, we design filter banks based on such tags of training images of the dataset which yields filter words (codebook) corresponding to the dataset.
Finally, for each input image which are represented by the tags, we design our proposed features as the histogram based on the codebook.

We evaluate the performance of the proposed features in scene image classification on three popular scene image datasets: MIT-67\cite{quattoni_recognizing_2009}, Scene15\cite{fei-fei_bayesian_2005} and Event8\cite{li2007and}.
 Our approach typically produces a smaller feature size than existing feature extraction methods. The experimental results suggest that the proposed features generate higher classification accuracies than vision-based and tag-based features and comparable classification to deep learning based features. 
 
\section{Related Works}
\label{sec:1}

Generally, there are three types of image features: (i) traditional vision-based features \cite{zeglazi_sift_2016,oliva2005gist,oliva_modeling_2001,dalal2005histograms,wu_centrist:_2011,xiao_mcentrist:_2014, margolin2014otc}, (ii) deep learning based features\cite{7968351,khan_discriminative_2016,8085139,guo_locally_2017,tang_g-ms2f:_2017}, and (iii) tag-based features\cite{zhang2017image,wang2019task}.

Traditional vision-based features are extracted based on the algorithms such as (SIFT)\cite{zeglazi_sift_2016}, Generalized
Search Trees (GIST)\cite{oliva2005gist},\cite{oliva_modeling_2001}, Histogram of Gradient (HOG)\cite{dalal2005histograms},
GIST-color\cite{oliva_modeling_2001}, 
SPM\cite{lazebnik2006beyond},
CENsus TRansform hISTogram (CENTRIST)\cite{wu_centrist:_2011}, multi-channel (mCENTRIST)\cite{xiao_mcentrist:_2014}, OTC\cite{margolin2014otc}, and so on. These features depend on the core information of the image such as colors, intensity, etc. Broadly, these features are computed in the local sense and are also called low-level features. These features are suitable for certain areas such as texture images. These features usually have high dimensions.

Similarly, deep learning based features, such as bilinear\cite{7968351}, Deep Un-structured Convolutional Activations (DUCA)\cite{khan_discriminative_2016}, Bag of Surrogate Parts (BoSP)\cite{8085139}, Locally Supervised Deep Hybrid Model (LS-DHM)\cite{guo_locally_2017} and GMS2F\cite{tang_g-ms2f:_2017}, are extracted from the intermediate layers of deep learning models (pre-trained models or user-defined models). Different layers of deep learning models provide different semantics information related to the input image. Thus, they have the capability to extract discriminating features compared to traditional vision based features. Deep learning based features enjoy prominent successes in image classification. 

Two prior works \cite{zhang2017image,wang2019task} have been recently presented for scene image recognition using tags-based features.
Zhang et al. \cite{zhang2017image} used the search engine to extract the description/annotation tags and designed Bag of Words (BoW) straightly. Their method ignores the concepts of filter banks and creates high-dimensional features which yield limited classification accuracy. Similarly, Wang et al. \cite{wang2019task} designed task-generic filters using pre-defined labels for the scene images but lacked task-specific filters to work in a specific domain. Also, because of the usage of pre-defined labels to design filter banks, the contextual information related to the images is hard to achieve. Moreover, due to the out of vocabulary (OOV) problem 
while constructing filter banks, their method  discards the tags that are not present in the WordNet\cite{miller1995wordnet},
and may contain insufficient vocabularies.

\begin{figure*}[t]
    \centering
    \includegraphics[width=0.90\textwidth, height=15cm,keepaspectratio]{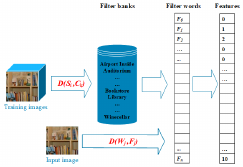}
    \caption{Overview of the proposed method. $D(S_i, C_k)$ denotes averaged similarity for tags $S_i$ and the category label $C_k$. $D(W_j,F_j)$ represents the semantic similarity of annotation/description tags $W_j$ with the filter words $F_j$. The filter words are generated by concatenating the filtered tags from the filter banks. Finally, based on the filter words (codebook), we design the histogram and accumulate the histograms for each bin of the features. The resulting features can semantically represent the input image. }
    \label{fig:3}
\end{figure*}

\section{The Proposed Method}
\label{sec:2}

Before using an actual pipeline of our proposed method, 
we first extract annotation/description tags of each image on the dataset using Yandex search engine\footnote{https://www.yandex.com/images/} where we select only top $k$ (i.e., $k=50$) visually similar images ranked by the search engine. As suggested in \cite{wang2019task,wang2019learning}, we use $k=50$ for the extraction of the tags. 
Our method consists of two major steps: (1) design of filter banks (Section \ref{sec:methodstep1}), and (2) feature extraction (Section \ref{sec:methodstep2}). The overall flow of our proposed method is shown in Fig. \ref{fig:3}. 

\subsection{Design of filter banks using training images}
\label{sec:methodstep1}
In this section, we describe how to design filter banks using the training set of one dataset.
We use training images to design the filter bank for each category.
To extract the filter words using filter banks, we present the following two steps. 

\subsubsection{Pre-processing of annotation/description tags}
\label{preprocessing}
After the extraction of annotation/description tags of the images under various categories, we pre-process them by removing punctuation marks, numbers, tokenization and language translation. Some of the extracted tags are also in the Russian language. We simply translate them into English using Google translator. Similarly, we remove the numeric content from the tags because numbers are not related to our purpose. We represent the tags of an image as ${\{W_{j}\}_{j=1}^m}$, 
where $m$ is the total number of tags in the image.

\subsubsection{Filter banks with semantic similarity}
We focus on task-specific filter banks which utilize the contextual information related to the image. We observe the fact that the tags of an image are semantically related to its category (or category label). Because raw tags for a training image are extracted from annotations of its $k=50$ most similar images in the web, the number of tags can be very long. To reduce the number of tags for the image, we select a subset of tags ($S_i \subset W_j$) which have more semantic similarity with the category label. First, we select the top $500$ frequent tags per training image of each category and then calculate their semantic similarity to the corresponding category label. We represent tags and category labels as two word embedding vectors
\cite{mikolov2013efficient,pennington2014glove,bojanowski2017enriching}
and use the cosine function 
to compute the semantic similarity (Eq. \eqref{eq:1}). 
\begin{equation}
\label{eq:1}
 cos(\pmb a, \pmb b ) = \frac {\pmb a \cdot \pmb b}{||\pmb a|| \cdot ||\pmb b||},
\end{equation} 
where $\pmb a$ and $\pmb b$ are the two embedding vectors.
For word embeddings, we utilize three popular pre-trained words embedding models -
Word2Vec\cite{mikolov2013efficient}, Glove\cite{pennington2014glove} and fastText\cite{bojanowski2017enriching}.
We find the final similarity by averaging the semantic similarity of tag and category label over the three types of word embedding vectors.
The averaging strategy usually helps to mitigate the OOV problem. Also, it exploits the knowledge of three domains from three embedding models.
To be efficient, we select only those tags whose averaged similarity ($D$) to the category label is greater than or equal to an empirical threshold of $\delta=0.50$.
\begin{equation}
\label{eq:3}
{S_i} =
\begin{cases} 
      1 & \text{if $D({S_i},{C_k}) \geq$ $\delta$}, \\
      0 & \text{otherwise}.
   \end{cases}
\end{equation}
We extract the filter bank for each category $C_k$ (the $k^{th}$ category) with the tags belonging to it.
From Eq. \eqref{eq:3}, we determine whether the particular tag $S_i$ is eligible to the filter bank for the corresponding category.  
The strategy to accept and reject the tags are represented by $1$ and $0$, respectively.

\begin{table}[t]
\caption{Sample filter banks of some categories in the MIT-67 dataset.}
\centering
\begin{tabular}{p{2.6cm} p{9cm}}
\toprule
Category & Filter banks\\
\midrule
Airport inside & airport, terminal, city, flight, hotel, flights, aviation\\
Library & library, books, libraries, archives, collections\\
Winecellar & wines, cellar, whiskey, winemaker, beverages, grapes, tastings\\
Subway & metro, subway, train, railway, transit, tram\\
\bottomrule
\end{tabular}
\label{tab:0}
\end{table} 

Table \ref{tab:0} lists the filter banks of the categories for the MIT-67\cite{quattoni_recognizing_2009} dataset. 
We design three separate sets
of filter banks for the MIT-67\cite{quattoni_recognizing_2009}, Scene15\cite{fei-fei_bayesian_2005} and Event8\cite{li2007and} datasets in the experiments. 
We design filter banks for MIT-67 using the training images of the dataset. On MIT-67 dataset, the total number of filter banks is $67$ where each contains filtered tags.
We combine all those filter banks to obtain $1254$-D filter words.
Furthermore, for Scene15 and Event8, we design filter banks over 10 sets on the corresponding dataset.
We utilize the corresponding training images of each set to separately design the filter banks and obtain filter words. 

 \subsection{Extraction of proposed tag-based semantic features for input image}
 \label{sec:methodstep2}
 We first utilize the annotation/description tags of each image to calculate the proposed features. These tags are preprocessed using Section \ref{preprocessing}.
 Then, all the filter banks are concatenated to form a list of filtered tags (or semantic tags), i.e., filter words (codebook). 
Inspired by the BoW approach \cite{huang2008similarity}, we design the histogram features of the input image using this codebook.
Our codebook is functionally similar to the codebook obtained by the clustering algorithm \cite{Hartigan:1975:CA:540298}.
However, our filter banks are based on contextual information and filter outliers significantly compared to the existing filter banks \cite{wang2019task}.
After that, we calculate the pairwise similarity of each filtered word with the pre-processed annotation/description tags of the input image. Denote the filter words by ${\{F_j\}_{j=1}^n}$. If we have $n$ unique filter words and the input image contains $m$ annotation/description tags, then the total similarity calculation is $n*m$. To calculate the similarity, we use the same scheme as that in designing the filter bank of each category.
\begin{equation}
\label{eq:4}
{H_j} =
\begin{cases} 
      1 & \text{if $D({W_j},{F_j}) \geq$ T}, \\
      0 & \text{otherwise}.
   \end{cases}
\end{equation}
$H_j$ in Eq. \eqref{eq:4} represents the histogram based on the filter words $F_j$. Here, $1$ represents the tag accepts the similarity with the filter word, and $0$ means the tag rejects similarity. We design the histogram by taking all the pre-processed tags of the input image.
For the bin of features corresponding to that filter word $F_j$, we count the number of tags which have acceptable similarity with $F_j$.

\begin{figure}[b]
    \centering
    \includegraphics[width=0.75\textwidth, height=7cm,keepaspectratio]{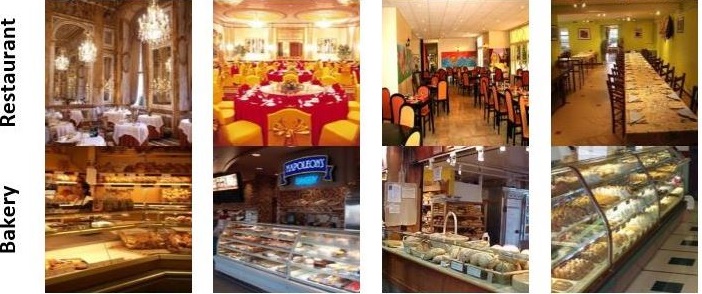}
    \caption{Sample images from the MIT-67 dataset.}
    \label{fig:4}
\end{figure}

\section{Experimental Results}
\label{sec:3}
In this section, we discuss the experimental setup and present results of our evaluation of the proposed features against other existing feature extraction methods in scene image classification using Support Vector Machine (SVM)\cite{Hearst:1998:SVM:630302.630387}.

\subsection{Implementation}
We use Yandex as the search engine and implement the proposed method using Python. 
We use the python SVM implementation available in the sklearn machine learning package\footnote{https://scikit-learn.org/stable/}.
We utilize the default setting for the majority of the parameters. However, we tune three parameters in the experiments.
The setting of two SVM parameters are: $kernel=rbf$ and $gamma=10^{-5}$. 
And, we tune the cost parameter $C$ in the range of $0$ to $100$ and tabulate only the setting that generates highest classification accuracy ($C=9$ for MIT-67 and $C=50$ for the remaining datasets). Furthermore, to take advantage of lightweight word-embedding vectors \cite{mikolov2013efficient, pennington2014glove, bojanowski2017enriching} of the tags, we use a popular Python package pymagnitude \cite{patel2018magnitude}. 
We fix a threshold of $0.50$ (i.e., $\delta=0.50$) to design the filter banks in Eq. \eqref{eq:3}
and empirically set a threshold of $0.40$ (i.e., $T=0.40$ in Eq. \eqref{eq:4})
for the extraction of our features which produces best results compared to other thresholds.
We will discuss about the threshold $T$ for the extraction of our features in Section \ref{threshold_analyze}.

\subsection{Datasets}
We use three publicly available datasets in our experiments: MIT-67\cite{quattoni_recognizing_2009}, Scene15\cite{fei-fei_bayesian_2005}, and Event8\cite{li2007and}. 
The MIT-67 dataset contains pictures of 67 categories. This is a challenging dataset which has noticeable intra-class and inter-class barriers (e.g., Fig. \ref{fig:4}). As defined in \cite{quattoni_recognizing_2009}, the number of training and testing images per category are $80$ and $20$, respectively.

Scene15 includes both indoor and outdoor images. It has $15$ categories. 
As with the previous works \cite{oliva_modeling_2001,
lazebnik2006beyond,
wu_centrist:_2011,
margolin2014otc,
lin_learning_2014,
zhang2017image,
tang_g-ms2f:_2017,
kim2014convolutional,
wang2019task}, we design $10$ sets of train/test split, where each split randomly contains $100$ images for training and remaining images for testing per category, and note the mean accuracy. 

Event8 involves images of $8$ sports categories. This dataset does not have pre-defined train/test splits, either. Like Scene15, we design $10$ sets of train/test split as in previous works \cite{li2010object, lin_learning_2014, shenghuagao2010local, perronnin2010improving, zhang2017image, kuzborskij2016naive, zhou2016places, he2016deep, wang2019task, kim2014convolutional} and note the mean accuracy.
For each split, we randomly select 130 images per category and divide 70 images for training and 60 images for testing.
\begin{table}[!hb] 
\caption{Comparisons of classification accuracy (\%) for the existing methods and ours on the three datasets.
The dash (-) symbol stands for no published accuracy on the specific dataset. }
\centering
\begin{tabular}{p{6.5cm} p{1.7cm} p{1.7cm} p{1.7cm}}
\toprule
Method & MIT-67 & Scene15 & Event8 \\
\midrule
Traditional computer vision-based methods\\
\midrule
GIST-color\cite{oliva_modeling_2001} & -&69.5 &- \\
ROI with GIST\cite{quattoni_recognizing_2009} &26.1 &- &-\\
SPM\cite{lazebnik2006beyond} &- &81.4 &- \\
MM-Scene\cite{zhu_large_2010} & 28.3 &- &-\\
CENTRIST\cite{wu_centrist:_2011}&- &83.9 &- \\
Object Bank\cite{li2010object}&37.6 &- &76.3\\
RBoW\cite{parizi2012reconfigurable}&37.9 &- &-\\
BOP\cite{juneja2013blocks}&46.1&- &-\\
OTC\cite{margolin2014otc}&47.3&84.4 &-\\
ISPR\cite{lin_learning_2014}&50.1&85.1 &74.9\\
LscSPM\cite{shenghuagao2010local}&- &- &85.3\\
IFV\cite{perronnin2010improving}&- &- &90.3\\
\midrule
Deep learning-based methods\\
\midrule
EISR \cite{zhang2017image} &66.2 &94.5 & 92.7 \\
CNN-MOP\cite{gong_multi-scale_2014} & 68.0&- &- \\
BoSP\cite{8085139} & 78.2&- &- \\
G-MS2F\cite{tang_g-ms2f:_2017}& 79.6&92.9 &-\\
CNN-sNBNL\cite{kuzborskij2016naive} &- &- &95.3 \\
VGG\cite{zhou2016places} &- &- &95.6 \\
ResNet152\cite{he2016deep} &- &- &96.9  \\
\midrule
Tag-based methods\\
\midrule
BoW\cite{wang2019task} &52.5&70.1 & 83.5 \\
CNN \cite{kim2014convolutional} & 52.0&72.2 &85.9\\
s-CNN(max)\cite{wang2019task} & 54.6&76.2 &90.9 \\
s-CNN(avg)\cite{wang2019task} & 55.1&76.7 &91.2\\
s-CNNC(max)\cite{wang2019task} & 55.9&77.2 &91.5\\
\textbf{Ours}& \textbf{76.5} & \textbf{81.3} &\textbf{94.4} \\
\hline
\end{tabular}
\label{tab:1}
\end{table} 

\subsection{Comparison with existing methods}
We compare the classification accuracy of our proposed features with the existing features which include traditional vision-based features, deep learning based features and tag-based features on the three datasets. The statistical accuracy numbers are listed in Table \ref{tab:1}. To minimize the bias, we compare our accuracy with the existing published accuracy on the same dataset. We straightforwardly take the results of existing features from corresponding papers. 

In the first column of Table \ref{tab:1}, we see that BoW yields an accuracy of $52.5\%$, which is the lowest accuracy among the tag-based methods. Researchers improve the accuracy of tag-based features using Convolutional Neural Network (CNN) model up to $55.9\%$\cite{wang2019task}. 
 We suspect these methods still could not provide highly discriminable features of the images. By contrast, deep learning based features improve classification accuracy. 
For example, BoSP, the deep learning based features, has over $4000$-D features size and its accuracy is higher than ours ($78.2\%$ versus $76.5\%$).
Our semantic features, which are based on annotation tags, provide prominent accuracy in image classification. We notice that our proposed features generate the highest accuracy of \textbf{76.5\%} among the tag-based methods.
Despite that the accuracy looks slightly lower than some of the deep learning based methods that benefit from the high-dimensional features, our features still outperform some of them \cite{zhang2017image,gong_multi-scale_2014,khan_discriminative_2016}.
Our method soundly outperforms the traditional vision based methods\cite{quattoni_recognizing_2009,zhu_large_2010,li2010object,parizi2012reconfigurable,juneja2013blocks,margolin2014otc,lin_learning_2014}. Our method leads to a noticeably smaller size of features 
on this dataset compared to other features (e.g., $1254$-D). 
The feature size differs, due to the number of categories and size of filter banks on different datasets.

The classification accuracies of different features on the Scene15 dataset are provided in the second column of Table \ref{tab:1}. Since our method belongs to the tag-based methods, we first compare our features against existing tag-based features. The BoW method provides an accuracy of $70.1\%$, which is the lowest among the tag-based methods on this dataset. With the use of CNN on tag-based methods,
the accuracy surges up to $77.2\%$. These methods, however, suffer from a large feature size. Since our features are dependent on the size of filter words ($<100$) on this dataset, our feature size is less than $100$ which is significantly lower than other features. Despite that, we observe that the proposed features have a prominent accuracy of $\textbf{81.3\%}$ among the tag-based features. 

In the third column of Table \ref{tab:1}, we enlist the classification accuracies of different features on the Event8 dataset. In this dataset, the BoW method provides an accuracy of $83.5\%$, the lowest accuracy among the tag-based methods. Moreover, by using CNN on tag-based methods, 
the accuracy increases up to $91.5\%$.  
Similarly, our features have a very low size ($<50$) on this dataset and are remarkably lower than other features. We achieve the best accuracy of $\textbf{94.4\%}$ among the tag-based features.

\subsection{Ablative study of threshold}
\label{threshold_analyze}
We analyze the effects of different thresholds $T$ in this section. 
To study the thresholds in depth, we use the Event8 dataset and follow the setup as above. We test thresholds between $0.30$ and $0.80$ with a step size of $0.10$. We summarize the classification accuracy of the proposed features with the corresponding thresholds in Table \ref{tab:4}. The best accuracy ($94.41\%$) is obtained by $T=0.40$, whereas the worst accuracy ($60.4\%$) is produced by $T=0.80$ on the dataset. We empirically observe that $0.40$ is a suitable threshold for all datasets, so we use it in all experiments. 

\begin{table}[t]
\caption{
Average accuracy over $10$ sets corresponding to different thresholds ($T$ in Eq. \eqref{eq:4}) on the Event8 dataset.
}
\centering
\begin{tabular}{p{2.2cm} p{1.5cm} p{1.5cm} p{1.5cm} p{1.5cm} p{1.5cm}p{1.5cm}}
\toprule
Threshold &0.30 &0.40 &0.50 &0.60  &0.70 &0.80\\
\midrule
Accuracy (\%) &93.7 & \textbf{94.4} & 93.0 &89.4 & 87.5&60.4 \\
\bottomrule
\end{tabular}
\label{tab:4}
\end{table}

\subsection{Ablative study of individual embedding}
In this section, we study the proposed features based on individual word embedding and the averaged semantic similarity scheme. We set the threshold $T=0.40$ and conduct experiments on the Event8 dataset and compute the mean accuracy over $10$ sets.

Table \ref{tab:5} shows the accuracies generated by our proposed features based on the individual embedding and the averaged semantic similarity. It seems that the features induced by the averaged semantic similarity produce a higher accuracy ($\textbf{94.4\%}$) than features of individual embeddings. This is because the features induced by the averaged semantic similarity act as the combined knowledge from three domains, which typically possess a higher separability than the individual embedding based features.

\begin{table}[t]
\caption{Accuracy of the proposed features using the individual embedding and averaged semantic similarity on the Event8 dataset.  }
\centering
\begin{tabular}{p{2.6cm} p{2.2cm} p{2.2cm} p{2.2cm} p{2.2cm}}
\toprule
Embeddings & Word2Vec & Glove & fastText& Averaged\\
\midrule
Accuracy (\%) &94.3 &93.5 &93.1& \textbf{94.4}\\
\bottomrule
\end{tabular}
\label{tab:5}
\end{table}

\section{Conclusion}
\label{sec:4}
In this paper, we propose a novel method to extract tag-based semantic features for the representation of scene images. We achieve this by performing two consecutive steps which are the design of filter banks and extraction of tag-based semantic features. We conduct experiments on three popular datasets and find that the proposed features produce better or comparable results to existing vision based, deep learning based and tag-based features, given a noticeably lower feature size of ours than those features.
In the future, we would like to investigate the incorporation of the proposed features and deep features to further improve image classification accuracy.

\bibliographystyle{splncs04}
\bibliography{Reference}

\begin{thebibliography}{10}
\providecommand{\url}[1]{\texttt{#1}}
\providecommand{\urlprefix}{URL }
\providecommand{\doi}[1]{https://doi.org/#1}

\bibitem{bojanowski2017enriching}
Bojanowski, P., Grave, E., Joulin, A., Mikolov, T.: Enriching word vectors with
  subword information. Transactions of the Association for Computational
  Linguistics  \textbf{5},  135--146 (2017)

\bibitem{dalal2005histograms}
Dalal, N., Triggs, B.: Histograms of oriented gradients for human detection.
  In: Proc. IEEE Comput. Soc. Conf. Comput. Vis. Pattern Recognit. (CVPR). pp.
  886--893 (2005)

\bibitem{deng_imagenet:_2009}
Deng, J., Dong, W., Socher, R., Li, L.J., Li, K., Fei-Fei, L.: {ImageNet}: {a}
  {large}-{scale} {hierarchical} {image} {database}. In: Proc. {IEEE} {Conf.}
  {Comput.} {Vis.} {Pattern} {Recognit.} ({CVPR}) (2009)

\bibitem{fei-fei_bayesian_2005}
Fei-Fei, L., Perona, P.: A bayesian hierarchical model for learning natural
  scene categories. In: Proc. {IEEE} {Comput.} {Soc.} {Conf.} {Comput.} {Vis.}
  and {Pattern} {Recognit.} ({CVPR}). vol.~2, pp. 524--531 (Jun 2005)

\bibitem{gong_multi-scale_2014}
Gong, Y., Wang, L., Guo, R., Lazebnik, S.: Multi-scale {orderless} {pooling} of
  {deep} {convolutional} {activation} {features}. In: Proc. Eur. Conf. Comput.
  Vis. (ECCV). pp. 392--407 (2014)

\bibitem{guo_locally_2017}
Guo, S., Huang, W., Wang, L., Qiao, Y.: Locally {supervised} {deep} {hybrid}
  {model} for {scene} {recognition}. {IEEE} Trans. Image Process.
  \textbf{26}(2),  808--820 (Feb 2017)

\bibitem{8085139}
Guo, Y., Liu, Y., Lao, S., Bakker, E.M., Bai, L., Lew, M.S.: Bag of surrogate
  parts feature for visual recognition. IEEE Trans. Multimedia  \textbf{20}(6),
   1525--1536 (Jun 2018)

\bibitem{Hartigan:1975:CA:540298}
Hartigan, J.A.: Clustering Algorithms. John Wiley \& Sons, Inc., New York, NY,
  USA, 99th edn. (1975)

\bibitem{he2016deep}
He, K., Zhang, X., Ren, S., Sun, J.: Deep residual learning for image
  recognition. In: Proc. IEEE Conf. on Computer Vision and Pattern Recognition
  (CVPR). pp. 770--778 (2016)

\bibitem{Hearst:1998:SVM:630302.630387}
Hearst, M.A.: Support vector machines. IEEE Intelligent Systems
  \textbf{13}(4),  18--28 (Jul 1998)

\bibitem{huang2008similarity}
Huang, A.: Similarity measures for text document clustering. In: Proc. Sixth
  New Zealand Computer Science Research Student Conference (NZCSRSC2008). pp.
  9--56 (2008)

\bibitem{juneja2013blocks}
Juneja, M., Vedaldi, A., Jawahar, C., Zisserman, A.: Blocks that shout:
  Distinctive parts for scene classification. In: Proc. IEEE Conf. Comput. Vis.
  Pattern Recognit. (CVPR). pp. 923--930 (Jun 2013)

\bibitem{khan_discriminative_2016}
Khan, S.H., Hayat, M., Bennamoun, M., Togneri, R., Sohel, F.A.: A
  {discriminative} {representation} of {convolutional} {features} for {indoor}
  {scene} {recognition}. {IEEE} Trans. Image Process.  \textbf{25}(7),
  3372--3383 (Jul 2016)

\bibitem{kim2014convolutional}
Kim, Y.: Convolutional neural networks for sentence classification. arXiv
  preprint arXiv:1408.5882  (2014)

\bibitem{kuzborskij2016naive}
Kuzborskij, I., Maria~Carlucci, F., Caputo, B.: When naive bayes nearest
  neighbors meet convolutional neural networks. In: Proc. IEEE Conf. on
  Computer Vision and Pattern Recognition (CVPR). pp. 2100--2109 (2016)

\bibitem{lazebnik2006beyond}
Lazebnik, S., Schmid, C., Ponce, J.: Beyond bags of features: Spatial pyramid
  matching for recognizing natural scene categories. In: Proc. IEEE Comput.
  Soc. Conf. Comput. Vis. Pattern Recognit. pp. 2169--2178 (Jun 2006)

\bibitem{li2007and}
Li, L.J., Li, F.F.: What, where and who? classifying events by scene and object
  recognition. In: ICCV. vol.~2, p.~6 (2007)

\bibitem{li2010object}
Li, L.J., Su, H., Fei-Fei, L., Xing, E.P.: Object bank: A high-level image
  representation for scene classification \& semantic feature sparsification.
  In: Proc. Adv. Neural Inf. Process. Syst. (NIPS). pp. 1378--1386 (2010)

\bibitem{lin_learning_2014}
Lin, D., Lu, C., Liao, R., Jia, J.: Learning {important} {spatial} {pooling}
  {regions} for {scene} {classification}. In: Proc. IEEE Conf. Comput. Vis.
  Pattern Recognit. (CVPR). pp. 3726--3733 (Jun 2014)

\bibitem{7968351}
Lin, T.Y., RoyChowdhury, A., Maji, S.: Bilinear convolutional neural networks
  for fine-grained visual recognition. {IEEE} Trans. Pattern Anal. Mach.
  Intell.  \textbf{40}(6),  1309--1322 (Jun 2018)

\bibitem{margolin2014otc}
Margolin, R., Zelnik-Manor, L., Tal, A.: Otc: A novel local descriptor for
  scene classification. In: Proc. Eur. Conf. Comput. Vis. (ECCV). pp. 377--391
  (2014)

\bibitem{mikolov2013efficient}
Mikolov, T., Chen, K., Corrado, G., Dean, J.: Efficient estimation of word
  representations in vector space. arXiv preprint arXiv:1301.3781  (2013)

\bibitem{miller1995wordnet}
Miller, G.A.: Wordnet: a lexical database for english. Commun. ACM
  \textbf{38}(11),  39--41 (1995)

\bibitem{oliva2005gist}
Oliva, A.: Gist of the scene. In: Neurobiology of Attention, pp. 251--256.
  Elsevier (2005)

\bibitem{oliva_modeling_2001}
Oliva, A., Torralba, A.: Modeling the {shape} of the {scene}: {a} {holistic}
  {representation} of the {spatial} {envelope}. Int. J. Comput. Vis.
  \textbf{42}(3),  145--175 (May 2001)

\bibitem{parizi2012reconfigurable}
Parizi, N., Oberlin, J.G., Felzenszwalb, P.F.: Reconfigurable models for scene
  recognition. In: Proc. Comput. Vis. Pattern Recognit.(CVPR). pp. 2775--2782
  (Jun 2012)

\bibitem{patel2018magnitude}
Patel, A., Sands, A., Callison-Burch, C., Apidianaki, M.: Magnitude: A fast,
  efficient universal vector embedding utility package. In: Proc. Conf. on
  Empirical Methods in Natural Language Processing: System Demonstrations. pp.
  120--126 (2018)

\bibitem{pennington2014glove}
Pennington, J., Socher, R., Manning, C.: Glove: Global vectors for word
  representation. In: Proc. 2014 Conf. on empirical methods in natural language
  processing (EMNLP). pp. 1532--1543 (2014)

\bibitem{perronnin2010improving}
Perronnin, F., S{\'a}nchez, J., Mensink, T.: Improving the fisher kernel for
  large-scale image classification. In: Proc. European Conference on Computer
  vision (ECCV). pp. 143--156 (2010)

\bibitem{quattoni_recognizing_2009}
Quattoni, A., Torralba, A.: Recognizing indoor scenes. In: Proc. {IEEE} {Conf.}
  {Comput.} {Vis.} {Pattern} {Recognit.} ({CVPR}). pp. 413--420 (Jun 2009)

\bibitem{shenghuagao2010local}
ShenghuaGao, I.H., Liang-TienChia, P.: Local features are not lonely--laplacian
  sparse coding for image classification pp. 3555--3561 (2010)

\bibitem{sitaula2019indoor}
Sitaula, C., Xiang, Y., Zhang, Y., Lu, X., Aryal, S.: Indoor image
  representation by high-level semantic features. IEEE Access  (Forthcoming)

\bibitem{tang_g-ms2f:_2017}
Tang, P., Wang, H., Kwong, S.: G-{MS2F}: {GoogLeNet} based multi-stage feature
  fusion of deep {CNN} for scene recognition. Neurocomputing  \textbf{225},
  188 -- 197 (Feb 2017)

\bibitem{wang2019task}
Wang, D., Mao, K.: Task-generic semantic convolutional neural network for web
  text-aided image classification. Neurocomputing  \textbf{329},  103--115
  (2019)

\bibitem{wang2019learning}
Wang, D., Mao, K.: Learning semantic text features for web text aided image
  classification. IEEE Trans. Multimedia  (2019)

\bibitem{wu_centrist:_2011}
Wu, J., Rehg, J.M.: {CENTRIST}: {A} {visual} {descriptor} for {scene}
  {categorization}. {IEEE} Trans. Pattern Anal. Mach. Intell.  \textbf{33}(8),
  1489--1501 (Aug 2011)

\bibitem{xiao_mcentrist:_2014}
Xiao, Y., Wu, J., Yuan, J.: {mCENTRIST}: {a} {multi}-{channel} {feature}
  {generation} {mechanism} for {scene} {categorization}. {IEEE} Trans. Image
  Process.  \textbf{23}(2),  823--836 (Feb 2014)

\bibitem{zeglazi_sift_2016}
Zeglazi, O., Amine, A., Rziza, M.: Sift {descriptors} {modeling} and
  {application} in {texture} {image} {classification}. In: Proc. 13th Int.
  Conf. Comput. Graphics, Imaging and Visualization (CGiV). pp. 265--268 (Mar
  2016)

\bibitem{zhang2017image}
Zhang, C., Zhu, G., Huang, Q., Tian, Q.: Image classification by search with
  explicitly and implicitly semantic representations. Information Sciences
  \textbf{376},  125--135 (2017)

\bibitem{zhou2016places}
Zhou, B., Khosla, A., Lapedriza, A., Torralba, A., Oliva, A.: Places: An image
  database for deep scene understanding. arXiv preprint arXiv:1610.02055
  (2016)

\bibitem{zhu_large_2010}
Zhu, J., Li, L.j., Fei-Fei, L., Xing, E.P.: Large {margin} {learning} of
  {upstream} {scene} {understanding} {models}. In: Proc. Adv. Neural Inf.
  Process. Syst. (NIPS), pp. 2586--2594 (2010)

\end{thebibliography}

\end{document}